\documentclass[letterpaper,10pt,conference]{ieeeconf}

\IEEEoverridecommandlockouts
 
\usepackage[english]{babel}
\usepackage[latin1]{inputenc}
\usepackage{graphicx}
\usepackage{amsmath,amssymb}
\usepackage{color}
\usepackage{subfigure}
\usepackage{url}
\usepackage{cite}
\usepackage{stfloats}
\usepackage{tikz}
\usepackage{balance}
\usepackage{gensymb}
\usepackage[algoruled,linesnumbered,noend]{algorithm2e}

\graphicspath{{./images/}}

\renewcommand{\textfloatsep}{0.2mm}
\renewcommand{\dbltextfloatsep}{0.2mm} 

\DeclareMathAlphabet\mathbfcal{OMS}{cmsy}{b}{n}

\newcommand{\x}{{x_1}}
\newcommand{\y}{{x_2}}

\newcommand{\agent}{$a_j$}

\newcommand{\agentSet}{$\textbf{a}$}

\newcommand{\prob}{$\textit{P}$}
\newcommand{\position}{$x^t_{a_j}$}

\newcommand{\velocity}{$v^t_{a_j}$}
\newcommand{\velocityPrev}{$v^{t-1}_{a_j}$}
\newcommand{\goalVelocity}{$v^t_{ji}$}

\newcommand{\goalVelocitySet}{$\textbf{v}^t_{ji}$}
\newcommand{\goal}{$g_i$}

\newcommand{\goalSet}{$\textbf{g}$}

\newcommand{\Planner}{$\mathcal{P}$}
\newcommand{\PlannerData}{$\mathcal{P}^{t}$}
\newcommand{\PlannerHist}{$\mathcal{P}^{t-1}$}
\newcommand{\Simulation}{$\mathcal{S}_{ji}$}
\newcommand{\SimulationSet}{$\mathbfcal{S}$}
\newcommand{\GoalPrior}{\prob(\goal)}
\newcommand{\Likelihood}{\prob(\velocity$|$\goal)}
\newcommand{\Posterior}{\prob(\goal$|$\velocity)}
\newcommand{\PosteriorPrev}{\prob(\goal$|$\velocityPrev)}
\newcommand{\GoalPriorM}{\textit{P}(g_i)}
\newcommand{\LikelihoodM}{\textit{P}(v^t_j|g_i)}
\newcommand{\PosteriorM}{\textit{P}(g_i|v^t_j)}

\newcommand{\BivGaussian}{\mathcal{N}_{\x,\y}}

\newcommand{\BivGaussianInst}{\BivGaussian\left(\boldsymbol{\mu}_{ji},\boldsymbol{\Sigma}_{j}\right)}
\newcommand{\mux}{\mu_\x}
\newcommand{\muy}{\mu_\y}

\title{\LARGE \bf Counterfactual Reasoning about Intent for\\ Interactive Navigation in Dynamic Environments}

\author{Alejandro Bordallo$^{1}$ \and Fabio Previtali$^{2}$ \and Nantas Nardelli$^{1}$ \and Subramanian Ramamoorthy$^{1}$
\thanks{\small $^{1}$Alejandro Bordallo, Nantas Nardelli and Subramanian Ramamoorthy are with the School of Informatics, University of Edinburgh, 10 Crichton Street, EH8 9AB, Edinburgh, Scotland}%
\thanks{\small $^{2}$Fabio Previtali is with the Department of Computer, Control, and Management Engineering, Sapienza University of Rome, via Ariosto 25, 00185, Rome, Italy}%
}

\begin{document}

\maketitle
\thispagestyle{empty}
\pagestyle{empty}

\begin{abstract}

\em Many modern robotics applications require robots to function autonomously in dynamic environments including other decision making agents, such as people or other robots. This calls for fast and scalable \textit{interactive} motion planning. This requires models that take into consideration the other agent's intended actions in one's own planning. We present a real-time motion planning framework that brings together a few key components including intention inference by reasoning counterfactually about potential motion of the other agents as they work towards different goals. By using a light-weight motion model, we achieve efficient iterative planning for fluid motion when avoiding pedestrians, in parallel with goal inference for longer range movement prediction. This inference framework is coupled with a novel distributed visual tracking method that provides reliable and robust models for the current belief-state of the monitored environment. This combined approach represents a computationally efficient alternative to previously studied policy learning methods that often require significant offline training or calibration and do not yet scale to densely populated environments. We validate this framework with experiments involving multi-robot and human-robot navigation. We further validate the tracker component separately on much larger scale unconstrained pedestrian data sets.

\end{abstract}

\section{Introduction}
\label{sec:Introduction}

Motion planning for mobile robotic platforms in human environments is a problem involving many constraints. Where and how the robot can travel is fundamentally defined by the environment and its evolution over time. For instance, the simplest motion planning specification is that the robot should not collide with entities in the environment. Given a model of the world, there are by now many standard approaches to computing trajectories that satisfy this simple requirement. However, the small modification that some entities in this environment can move around, on their own accord and possibly with their own separate goals, can have a substantial influence on the nature of the motion planning problem. Of the few methods that can cope with such dynamic environments, many depend on having access to significant amounts of prior knowledge (e.g., corpora of example movements from past experience) so as to train models of the dynamics of the environment which are then used for decision making. A standard approach, for instance, is to pose the problem in decision theoretic terms (e.g., using Partially Observable Markov Decision Processes or its variants), learning the necessary components of models from past data. However, this can be cumbersome in many application scenarios. Realistic navigation in crowded spaces is an intrinsically \textit{interactive} planning problem, which significantly increases the complexity of decision-theoretic formulations. Also, we often want robots to be deployable in multiple environments, which further stretches these methods in terms of model complexity and data requirements. So, on platforms that have resource constraints, there is an unmet need for efficient solutions to these interactive motion planning problems. 

We adopt an intermediate stance wherein we utilise a simple parameterised motion model (based on the concept of Hybrid Reciprocal Velocity Obstacles) that captures key elements of how people navigate when encountering other people in the same space; estimating the parameters of such a model from data. Our model is simple enough, structurally, to enable tractable learning from data. At the same time, it provides sufficient bias to incorporate what is otherwise often learnt in an expensive way from historical data. Furthermore, we utilise a tractable set of such models to define a belief-update computation over goals.

\begin{figure}[!t]
	\centering
	\begin{tikzpicture}[map/.style={draw=white,ultra thick,inner sep=0pt}]
		\node at (0,0) [map]
		{
			\includegraphics[width=\linewidth]{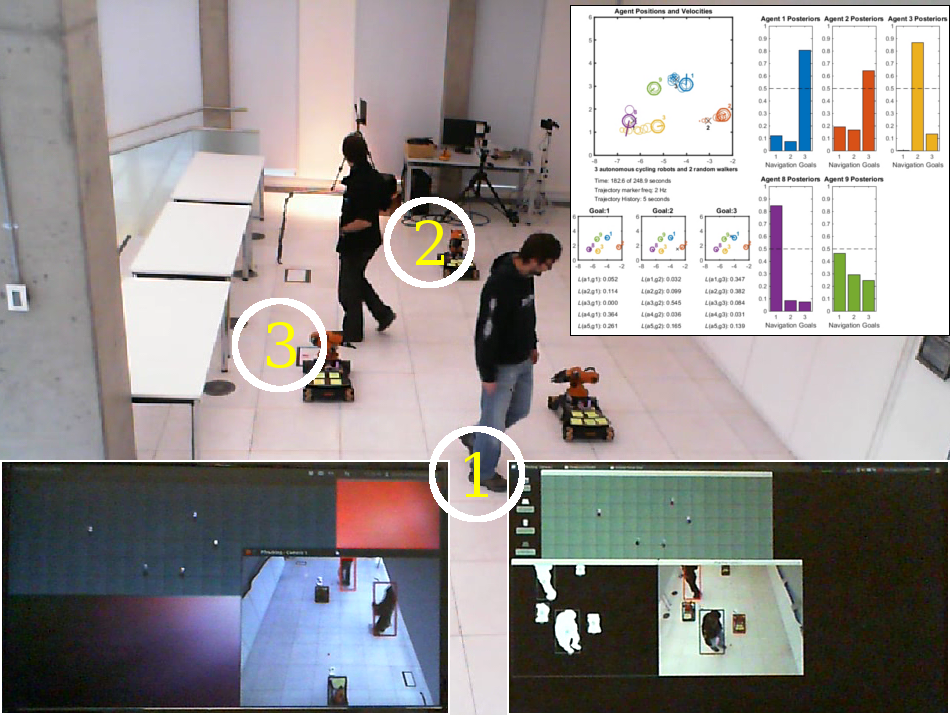}
		};
	\end{tikzpicture}
	\vspace{-0.7cm}
	\caption{Inferring intentions of three KUKA YouBots and two people by using our
			 counterfactual intention inference algorithm. Interactions among agents are
			 forced due to a limited collision-free navigation space. A novel distributed
			 tracking method is used to provide real-time motion data.}
	\label{fig:PersonYoubotScreen1}
\end{figure}

In our framework, we conceptualise each other agent as adopting locally-optimal actions given a potential goal. These goals, which represent movement intention, are of course latent and unobserved by our planning agent. So, the problem of said agent is to infer from noisy data these goals in real-time, enabling a trajectory to be planned over a longer horizon than reactive avoidance would. Intention-awareness is achieved by \textit{counterfactual} reasoning, using the predictions of the locally-optimal movement model to update beliefs regarding latent goals. The key contributions of this proposed framework are:

\begin{itemize}
	\item An intention-inference algorithm for dynamic environments with multiple
		  interactively navigating agents;
	\item A novel multi-camera multi-object tracking system, light weight yet flexible
		  enough to accommodate dynamically varying numbers of objects;
	\item An asynchronous distributed architecture to improve efficiency and robustness
		  (e.g. with respect to communication failures).
\end{itemize}
\noindent
We report on experiments with simulated and physical experiments in which robotic and human agents navigate autonomously, moving toward goals while naturally avoiding each other (see \figurename~\ref{fig:PersonYoubotScreen1}). Our robot planner runs robustly at 10Hz, navigating naturally around other agents - implicitly inferring the target goal of other agents in real-time using our inference model.
\section{Related Work}
\label{sec:RelatedWork}
 
\textbf{Interactive Motion Planning.} One could summarise progress in this domain by placing prior work in two major categories.

The first category, involving optimal planning, includes works that attempt to generatively describe various elements that influence human navigation behaviours, such as environment context-dependent navigation \cite{Ziebart09} or interacting social forces between agents\cite{Luber10}. While these approaches are often successful in achieving faithful description, they can also be computationally expensive. Moreover, it has been observed that attempting to achieve tractability in such models by shortening planning horizons can lead to pathological behaviour, such as `freezing' \cite{Trautman10} where no path seems feasible when one allows for potential evolution of uncertainty models.

This perhaps explains the popularity of simpler models which is the second category, e.g., constant velocity models for pedestrians combined with A* planning on a road-map environment \cite{Kummerle13}. Although these simpler methods do work in many large outdoor spaces, they can perform poorly when pedestrian density increases (e.g., crowded indoor environments). Working from this direction, to overcome the limitations of these simplistic motion models, it has been shown that offline training from demonstration data can yield optimal navigation policies and human-like trajectories, e.g., \cite{Kretzschmar14}. However, by the time we lift this to highly dynamic environments, e.g., \cite{Liu14}, the data requirements can become a burden. An alternative approach is to not model the environment iteratively but instead to derive an optimal policy from a navigation model and and fit its parameters online given the observed behaviour \cite{Bera14}. However, this can easily become suboptimal when the environment changes sufficiently.

\textbf{Multi-Object Tracking.} The problem of multiple object tracking has been addressed by many researchers, yielding many solutions each specialising the proposed approach to a chosen application field. Multi-object tracking algorithms can be classified in two groups: \emph{global} and \emph{recursive} \cite{Andriyenko11}.

Global (or \textit{offline}) methods formulate the tracking problem as one of optimisation, where all the trajectories within a temporal window are optimised jointly (e.g., \cite{Berclaz11,Ess09,Zhang08}). To be computationally tractable, such approaches try to restrict the space of possible object locations to a relatively small set of discrete points, either by first locating objects in each frame and then linking them together, or by using a discrete location grid. Berclaz \emph{et al.} \cite{Berclaz11} introduce a generic and mathematically sound multiple object tracking framework based on a $k$-shortest paths optimisation algorithm. Firstly, objects are detected in individual frames and then linked across frames allowing them to be very robust to false detections. Leal-Taix{\'e} \emph{et al.} \cite{Leal11} formulate a new graph model for the multiple object tracking challenge by minimising network flow. Another global approach, by Sharma \emph{et al.} \cite{Sharma09}, involves a Cluster-Boosted-Tree based pedestrian detector, adapted to deal with people tracking. The hierarchical association framework of Sharma \emph{et al.} first generates initial object tracklets by directly linking detection responses in neighbouring frames, and then progressively associates these tracklets to obtain final object tracks at multiple levels. However, the aforementioned methods allow the possibility of getting information from the future - a physically unrealistic feature that renders the methods only suitable to offline use.

On the other hand, recursive (or \textit{online}) methods estimate the current state relying only on the current observations and on the previous state. Early examples of such methods are Kalman filter based approaches (e.g., \cite{Makarenko04,Black02}), while more recent work usually uses particle filtering, allowing modelling of non-linear behaviours and multi-modal posterior distributions (e.g., \cite{Breitenstein11,Breitenstein09}). Breitenstein \emph{et al.} \cite{Breitenstein11} propose an online method for multi-person tracking-by-detection in a particle filtering framework obtaining good results. However, the designed approach cannot perform in real-time due to the low frame rate. Yang \emph{et al.} \cite{Yang09}, instead, design a probabilistic appearance model method to track multiple people through complex situations. Both the background and foreground models are described using Gaussian appearance models. However, the well-engineered system of Yang \emph{et al.} is not real-time and it relies on a static background, making the entire method weak when it comes to changes in the environment.
\section{Modelling Approach}
\label{sec:ModellingApproach}

\subsection{Intention Inference}

\begin{figure*}[!htb]
	\centering
	\subfigure[]
	{
		\begin{tikzpicture}[map/.style={draw=white,ultra thick,inner sep=0pt}]
			\node at (0,0) [map]
			{
				\includegraphics[height=3.45cm]{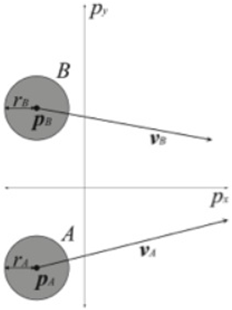}
			};
            \label{fig:HRVOevolution_a}
		\end{tikzpicture}
	}
	\hspace{0.6mm}
	\subfigure[]
	{
		\begin{tikzpicture}[map/.style={draw=white,ultra thick,inner sep=0pt}]
			\node at (0,0) [map]
			{
				\includegraphics[height=3.45cm]{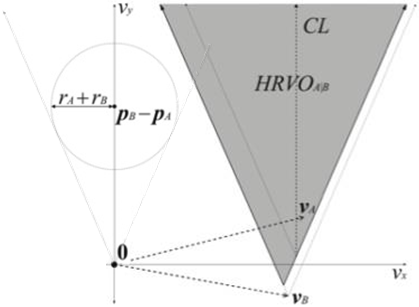}
			};
            \label{fig:HRVOevolution_b}
		\end{tikzpicture}
	}
	\hspace{0.6mm}
	\subfigure[]
	{
		\begin{tikzpicture}[map/.style={draw=white,ultra thick,inner sep=0pt}]
			\node at (0,0) [map]
			{
				\includegraphics[height=3.45cm]{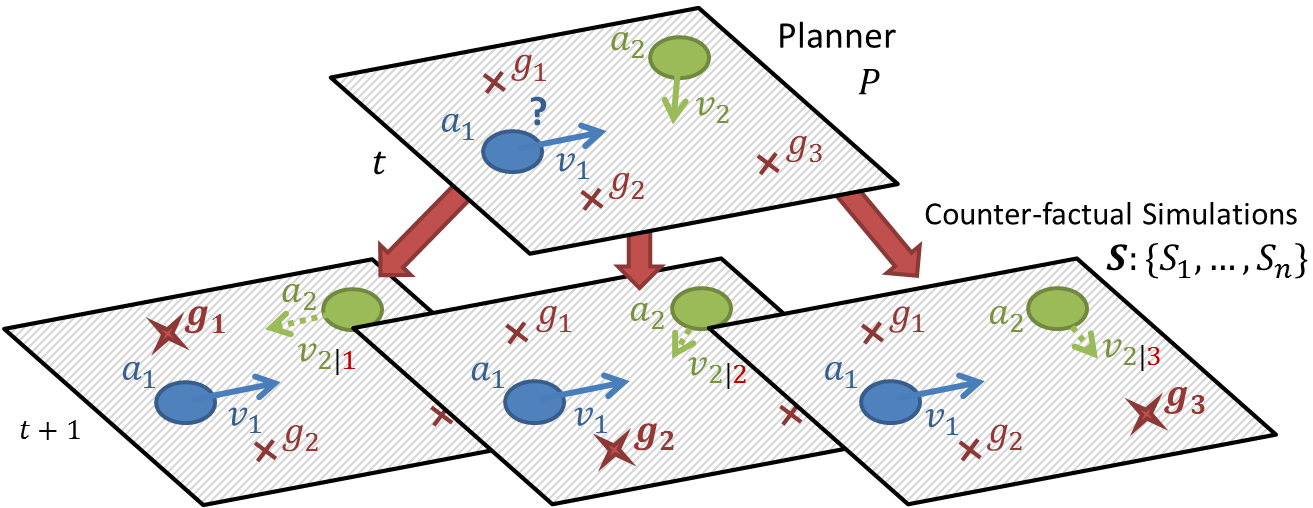}
			};
            \label{fig:HRVOevolution_c}
		\end{tikzpicture}
	}
	\vspace{-0.15cm}
    \caption{\textbf{Left}: two autonomous agents \agent\, navigating with instantaneous
    		 velocities \velocity. \textbf{Center}: Hybrid Velocity Obstacle $HRVO_{A|B}$
    		 for agent A influenced by agent B. \textbf{Right}: our counterfactual
    		 framework iteratively generates a set of simulated environments for each
    		 agent in the real world. Each simulation computes the locally optimal motion
    		 given each possible target goal. These velocities are compared with the
    		 observed agent motion using Bayesian recursive estimation for intention
    		 inference. \figurename~\ref{fig:HRVOevolution_a} and
    		 \ref{fig:HRVOevolution_b} are borrowed from Snape \emph{et al.}
    		 \cite{Snape13}.}
	\label{fig:HRVOevolution}
\end{figure*}

For each agent, \agent\,$\in$\,\agentSet\, that is detected and tracked in the environment, we compute predictions of movement intention in real-time. The `intention' of an agent is defined as the target goal, \goal\, that agent, \agent\,  is attempting to reach. The action space is defined as the set of possible velocities achievable in the next planning step given the agent's dynamic constraints. We construct the agent motion model by online parameter fitting given a stream of observed behavioural data, provided by the aforementioned distributed tracker (see Section~\ref{sec:PTracking}).

We then use these models to generate a set of plausible actions \goalVelocitySet\, where each \goalVelocity\, is the simulated locally optimal motion of \agent\, navigating towards \goal. These simulated velocity vectors \goalVelocity\, provide the motion probabilities required for estimation of the likelihood of \agent\, navigating to \goal, given the observed agent motion \velocity.

\subsection{Interactive Multi-Agent Navigation Framework}

Our parametrised interactive dynamics model is constructed based on the notion of Hybrid Reciprocal Velocity Obstacles (HRVO) \cite{Snape11}. Multi-agent simulators utilising this concept represent an efficient framework for simulating large numbers of agents navigating towards predefined goals while avoiding collisions with each other. These simulation runs iteratively, where in each time step all agents compute a new velocity vector. Their planned motion is constrained by the movements and positions of other agents, represented as velocity obstacles. The selected \textit{new velocity} is the closest to the \textit{preferred velocity}, the best unconstrained velocity towards the goal belonging to the subset of non-colliding velocities.
 
Originally designed for massive multi-agent simulations, the simulation maximises computation speed and scalability at the cost of short-sighted motion and agent collisions \cite{Kim14}. We utilise its advantages to perform fast deterministic sampling of agent motions for parameter fitting for densely populated indoor environments.

This motion model is inherently interactive, by considering the relationships between velocity obstacles implied my multiple agents, which enables our inference algorithm to usefully differentiate between purposeful advancement towards a goal and avoidance behaviours which could be mistaken as such. The framework is comparable to a constant velocity model whenever an agent is unobstructed.
\section{Goal inference algorithm}
\label{sec:GoalInference}

In our framework, we consider each agent to be pursuing a goal while avoiding collisions and minimising travel time. Each agent has an internal model of the environment and agents within it. In the context of such internal models, we consider our agents to be boundedly rational.

Each planning agent first performs a sensing update of all agent positions and velocities. Using the updated agent motion models, the planner agent infers the target goal of all agents given past observations. Finally, the planner computes a collision free motion given the inferred next movement of surrounding agents. We assume every other agent performs a similar but not necessarily identical procedure for navigating through the environment.
 
We now present the goal inference (Algorithm~\ref{alg:GoalInference}), which calculates the posterior distribution over possible goal intentions using Bayesian Recursive Estimation (Eq.~\ref{eq:GlobalLikelihood-1}).

\begin{algorithm}[!t]
	\small
	\caption{Goal Inference}\label{alg:GoalInference}
	
	\KwIn{set of goals \goalSet, agents to be modelled \agentSet, planner environment \Planner, inference history stored in \PlannerHist}
	\BlankLine
	\KwData{simulation environments \SimulationSet, simulated velocities \goalVelocitySet}
	\BlankLine
	\KwOut{updated intention posteriors \Posterior}
	\BlankLine
	\emph{Sensor update} \PlannerData $\rightarrow$ \Planner$:$\{\position, \velocity\} $ \; \forall $ \agent $ \in $ \agentSet  \\
	\ForEach{\agent $ \in $ \agentSet}
	{
		\BlankLine 
		\ForEach{\goal $ \in $ \goalSet}
		{
			\BlankLine 
			\emph{Instantiate} \Simulation $ \leftarrow $ \PlannerHist \\
			\BlankLine
			\emph{Set} \agent\,goal $ \leftarrow $ \goal \\
			\BlankLine
			\Simulation\, \textbf{do simulation step} \\
			\BlankLine
			\emph{Obtain} \goalVelocity $\leftarrow$ \Simulation \\
			\BlankLine
			$\BivGaussianInst$, Eq.~\ref{eq:BivariateGaussianDef} \\
			\BlankLine
			\Likelihood\, from $\BivGaussianInst$, Eq.~\ref{eq:BivariateGaussianDef} \\
			\BlankLine
			\If{\GoalPrior\,\emph{not initialised}}
			{
				\GoalPrior$=\frac{1}{\|\bf{g}\|} $
			}
			\BlankLine
			\emph{Update} \Posterior, Eq.~\ref{eq:PosteriorUpdate}
		}
    }
	\BlankLine 
	\emph{Return}\,\Posterior\, $\forall$\,\agent$,$\goal
\end{algorithm}

\textbf{Description.} The set of navigation goals \goalSet\, is provided \textit{a priori} (such as could be given by a semantic map). The goals represent the set of hypothetical intentions the planning agent \Planner\, considers for each agent \agent. Observed positions and velocities \position, \velocity\ $\forall$ \agentSet\ are updated during the sensing step and stored in \PlannerData. We then generate a simulation \Simulation\, of the environment for each \agent and \goal\,, transferring the up-to-date information of all agents to each instantiated \Simulation. Each simulated environment is run for a single time step, producing simulated \goalVelocity\, for each agent given the specified target goals. These velocities are constrained by \velocityPrev\, and \agent\, navigation parameters (average, maximum velocities and accelerations), which are updated online given sensor observations and stored on the planner agent's memory. See \figurename~\ref{fig:HRVOevolution_c} for a visual depiction of this process.

The set of simulated velocities \goalVelocitySet is used for generating the set of counterfactual motion probability distributions used by the inference algorithm. The posterior update rule for Bayesian Recursive Estimation is described as:
\begin{equation}
	\PosteriorM=\LikelihoodM\GoalPriorM
	\label{eq:PosteriorUpdate}
\end{equation}
\noindent
where \Posterior\, is the probability that agent \agent\, with current velocity \velocity\, is heading towards goal \goal. \GoalPrior\, is the prior probability for each \goal, initially uniformly distributed across all goals and updated after every inference step with the previously calculated posterior \PosteriorPrev. The likelihood \Likelihood\, of \velocity\, given \goal\, is sampled from a bivariate normal probability distribution constructed from each \goalVelocity\, such that:
\begin{equation}
	\BivGaussianInst\mbox{ , }\,\boldsymbol{\mu}_{ji}=\left(\begin{array}{c}
	\mux \\
	\muy\end{array}\right)
	\label{eq:BivariateGaussianDef}
\end{equation}
\noindent
where {$\boldsymbol{\mu}_{ji}$} is the mean for the bivariate gaussian distribution for \agent\, and \goal\, centered at \goalVelocity\,, or \Likelihood\, in  Eq.~\ref{eq:PosteriorUpdate}. After each iteration of the inference algorithm, the set of normalised posterior probabilities converges towards the latent intention of the agent. The most probable goal is then used by the planner agent to accurately predict the future motion of each \agent.

\begin{figure}
	\centering
	\includegraphics[width=0.95\linewidth]{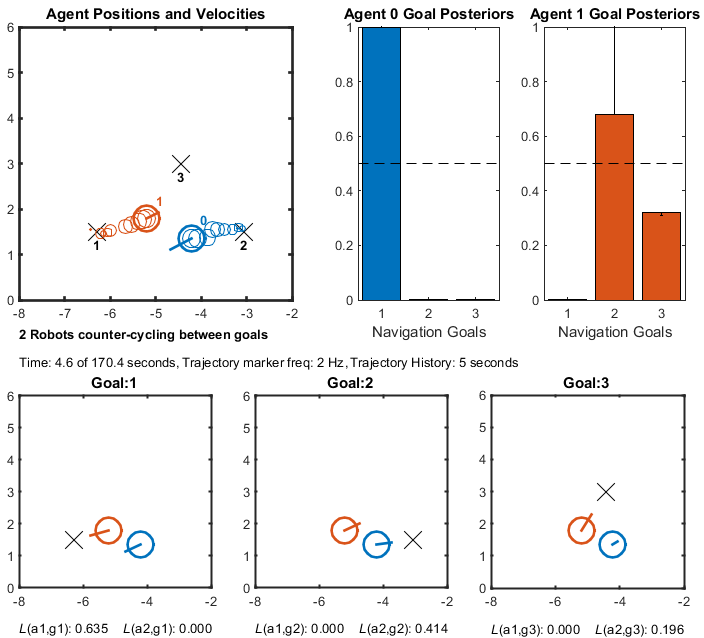}
	\vspace{-0.2cm}
	\caption{Two autonomous planning robots moving towards opposite goals. \emph{Agent1}'s bearing and velocity indicate movement towards \emph{Goal3}, but our inference algorithm correctly predicts its
             true intention towards \emph{Goal2}. Agent trails represent past trajectories, instantaneous likelihoods are shown under each counterfactual simulation window.} 
	\label{fig:Face2Face_1}
\end{figure}

As an example, consider two agents navigating autonomously between \emph{Goals 1} and \emph{2}, as seen in \figurename~\ref{fig:Face2Face_1}. The intersection between goals forces agents to evade each other while navigating towards their target. The velocity of \emph{Agent1} is, in an unobstructed scenario, closer to the optimal velocity towards \emph{Goal3} rather than \emph{2}. However, the presence and behaviour of \emph{Agent0} constraints the range of possible motions by \emph{Agent1} and vice versa. Our inference framework considers this and generates a set of counterfactual velocities for each agent given all possible goals and other agents present in the environment. So {$\textit{P}(v^t_1|g_2)>\textit{P}(v^t_1|g_3)$} and thus {$\textit{P}(g_2|v^t_1)$} increases towards iterative convergence.
\section{Distributed Multi-Camera Multiple\\Object Tracking}
\label{sec:PTracking}

\subsection{Problem Definition}
\label{sec:PTracking-PD}

The Distributed Multi-Camera Multiple Object Tracking problem can be formalised as follows. Let $ \mathcal{O} = \{ o_1, \dots, o_n \} $ be the set of all moving objects, each one having a different identity, and $ \mathcal{S} = \{s_1, \ldots, s_S\} $ be the set of arbitrarily fixed sensors, each one having limited knowledge about the environment (i.e., each camera can monitor only part of the scene). Moving objects are detected by a background subtraction algorithm and the number of objects $ n $ is unknown and can change over time. The set of measurements about the objects in the field-of-view of a camera $ s \in \mathcal{S} $ at a time $ t $ is denoted by $ z_{s,t} = \{ z_{s,t}^{(1)}, \ldots, z_{s,t}^{(l)} \} $, where a measurement $ z_{s,t}^{(i)} $ can be either a real object present in the environment or a false positive. The set of all the measurements gathered by all cameras at time $ t $ is denoted by $ z_{\mathcal{S},t} = \{z_{s,t} \, | \, s \in \mathcal{S}\} $. The history in time of all the measurements coming from all cameras is defined as $ z_{\mathcal{S},{1:t}} = \{z_{\mathcal{S},j} : 1 \leq j \leq t\} $. It is worth noticing that, we do not assume the measurements generated by the cameras to be synchronised. The goal is to determine, for each camera $ s $, an estimation $ x_{s,t} $ of the position of the objects at time $ t $ in a distributed fashion.

\subsection{Distributed Multi-Clustered Particle Filtering}

In order to achieve this goal, we estimate, for each camera $ s $, the position $ x_{s,t} = \{ x_{s,t}^{(1)}, \ldots, x_{s,t}^{(v)} \} $ of the objects by merging all the available information. Although the cameras continuously send information about their observations, the estimation computed by one camera may be different from the others due to noise or delay in communication. Specifically, the overall objective is to determine the likelihood $ p(x_{s,t} \, | \, z_{\mathcal{S},{1:t}}) $ of the global estimation $ x_{s,t} $ for each camera $ s $, given the observations $ z_{\mathcal{S},{1:t}} $ collected by all cameras.

We assume that the acquired observations are affected by an unknown noise that is conditionally independent among the cameras. During the acquisition process, each camera does not interact with the others, thus allowing for a factorisation of the likelihood of the global estimation that can be expressed by the following joint likelihood:
\begin{equation}
	\small
	p(z_{\mathcal{S},t} | x_{s,t}) = \prod\nolimits_{s \in \mathcal{S}} p(z_{s,t} | x_{s,t}) \label{eq:Factorization}
\end{equation}
\noindent
Given the assumption in Eq. (\ref{eq:Factorization}), a fusion algorithm can be described using Bayesian Recursive Estimation:
\small
\begin{eqnarray}
	p(x_{s,t} | z_{\mathcal{S},{1:t}}) = \frac{p(z_{\mathcal{S},t} | x_{s,t}) p(x_{s,t} | z_{\mathcal{S},{1:t-1}})}{\int p(z_{\mathcal{S},t} | x_{s,t})p(x_{s,t} | z_{\mathcal{S},{1:t-1}})dx_{s,t}} \;\;\;\;\;\;\;\;\;\;\;
	\label{eq:GlobalLikelihood-1}\\
	p(x_{s,t} | z_{\mathcal{S},{1:t-1}}) = \int p(x_{s,t} | x_{s,{t-1}}) p(x_{s,{t-1}} | z_{\mathcal{S},{1:t-1}})dx_{s,{t-1}}
	\label{eq:GlobalLikelihood-2}
\end{eqnarray}
\noindent
\normalsize
Eq. (\ref{eq:GlobalLikelihood-1}) and (\ref{eq:GlobalLikelihood-2}) represent a global recursive update that can be computed if and only if complete knowledge about the environment is available. Therefore, we propose to approximate the exact optimal Bayesian computation - Eq. (\ref{eq:GlobalLikelihood-1}) and (\ref{eq:GlobalLikelihood-2}) - by using a Distributed Particle Filter-based algorithm. To this end, we devise a novel method, called \emph{PTracking}, based on Distributed Multi-Clustered Particle Filtering. The algorithm is divided into two phases, namely a \emph{local estimation} phase and a \emph{global estimation} phase (Algorithm \ref{alg:MCPF}). Each camera performs the local and global computation, sharing the obtained results in order to achieve a better representation of the current scene.

The novelty of the proposed approach is in the integration of the following three main features: 1) a new clustering technique that keeps track of a variable unknown number of objects ensuring a limited distribution in the space of the particles; 2) the approximation of the particle distribution as \emph{Gaussian Mixture Models} (\emph{GMM}) to improve robustness and reduce the network overload; 3) an asynchronous approach to improve the flexibility and the robustness of the entire system (e.g., robustness to communication failures, dead nodes and so on).

\begin{algorithm}[!t]
	\small
	\caption{PTracking}\label{alg:MCPF}
	
	\KwIn{perceptions $ z_{s,t} $, local track numbers $ i_{s,t-1} $, global track numbers $ I_{s,t-1} $}
	\BlankLine
	\KwData{set of local particles $ \tilde{\xi}_{s,t} $, set of global particles $ \tilde{\xi}_{\mathcal{S'},t} $, local GMM set $ \mathcal{L} $, global GMM set $ \mathcal{G} $}
	\BlankLine
	\KwOut{global estimations $ x_{s,t} = (\boldsymbol{I}_{s,t},\boldsymbol\Lambda_{s,t},\boldsymbol{M}_{s,t},\boldsymbol\Sigma_{s,t}) $}
	\BlankLine
	\Begin
	{
		\label{lab:BeginLocalEstimation}
		$ \tilde{\xi}_{s,t} \sim \pi_t (x_{s,t} | x_{s,t-1},z_{s,t}) $
		\BlankLine
		Re-sample by using the SIR principle\\
		\BlankLine
		$ \mathcal{L} = KClusterize(\tilde{\xi}_{s,t}) $
		\BlankLine
		$ (\boldsymbol{i}_{s,t},\boldsymbol\lambda_{s,t},\boldsymbol\mu_{s,t},\boldsymbol\sigma_{s,t}) = DataAssociation(\mathcal{L}, i_{s,t-1}) $
		\BlankLine
		Communicate belief $ (\boldsymbol{i}_{s,t},\boldsymbol\lambda_{s,t},\boldsymbol\mu_{s,t},\boldsymbol\sigma_{s,t}) $ to other agents
	}
    \textbf{end}
	\label{lab:EndLocalEstimation}
	\BlankLine
	\Begin
	{
		\label{lab:BeginGlobalEstimation}
		Collect $ \mathcal{L}_{S'} $ from a subset $ \mathcal{S'} \subseteq \mathcal{S} $ of cameras within a $ \Delta t $
		\BlankLine
		$ \tilde{\xi}_{\mathcal{S'},t} \sim \tilde\pi = \sum_{s \in \mathcal{S'}} \boldsymbol\lambda_{s,t} \, \mathcal{N} (\boldsymbol\mu_{s,t},\boldsymbol\sigma_{s,t}) $
		\BlankLine
		Re-sample by using the SIR principle\\
		\BlankLine
		$ \mathcal{G} = KClusterize(\tilde\xi_{{\mathcal{S'},t}}) $
		\BlankLine
		$ (\boldsymbol{I}_{s,t},\boldsymbol\Lambda_{s,t},\boldsymbol{M}_{s,t},\boldsymbol\Sigma_{s,t}) = DataAssociation(\mathcal{G},I_{s,t-1}) $
	}
    \textbf{end}
	\label{lab:EndGlobalEstimation}
\end{algorithm}

\textbf{Local estimation.} The local estimation phase (Algorithm \ref{alg:MCPF}, lines \ref{lab:BeginLocalEstimation}-\ref{lab:EndLocalEstimation}) contains three steps: 1) A particle filtering step, that computes the evolution of the local estimations given the local observations $ z_{s,t} $ provided by the sensor; 2) A clustering step that determines the GMM parameters of this distribution; 3) A \emph{data association} step to assign an identity to each object $ o \in \mathcal{O} $.

The prediction step of the PF uses an initial guessed distribution, based on a \emph{transition state} model $ \pi $. Such a transition model makes a prediction of the next state based on the sensor movement. Then, using the previously computed state $ x_{s,{t-1}} $, the transition model, given by the measurements $ z_{s,t} $, is applied. Afterwards, from this hypothesised distribution, a set of samples is drawn and weighted exploiting the current local perception $ z_{s,t} $. Finally, the \emph{Sampling Importance Re-sampling} (\emph{SIR}) principle is used to re-sample the particles which are then clustered in order to determine the parameters of the final GMM model. It is worth noticing that, in contrast to other related approaches, this step enables the creation of a more compact information structure allowing us to drastically reduce the communication overhead. A \emph{data association} step is then applied to assign an identity (track number) to each object.

When the final GMM set has been computed, each camera broadcasts the set of GMM parameters describing all the objects detected.

\textbf{KClusterize.} The clustering phase is performed by using a novel clustering algorithm, called \emph{KClusterize}, aiming at fulfilling the following three requirements: 1) number of objects to be detected cannot be known a priori, 2) low computational load for real-time applications and 3) Gaussian distribution for each cluster. Alternative clustering methods are not adequate since they either need to know the number of clusters in advance (e.g., \emph{k-means}), or they are computationally expensive and not real-time (e.g., free-clustering algorithms like \emph{Expectation-Maximization}, \emph{BSAS} or \emph{QT-Clustering}). KClusterize does not require any initialisation, it has a linear complexity and all the obtained clusters reflect a Gaussian distribution.

\begin{table*}[!t]
	\centering
	\caption{Quantitative comparison on PETS 2009 with state-of-the-art methods.				 Results taken from corresponding paper of the authors.}
	\vspace{-0.2cm}
	\normalsize
	\begin{tabular}{ c | c | c | c | c | c | c | c | }
		\cline{2-8}
		& \multicolumn{1}{|c|}{\begin{tabular}[c]{@{}c@{}}Leal-Taix\'{e}\\ et al.\cite{Leal11} \end{tabular}} & \multicolumn{1}{|c|}{\begin{tabular}[c]{@{}c@{}}Berclaz\\ et al.\cite{Berclaz11} \end{tabular}} & \multicolumn{1}{|c|}{\begin{tabular}[c]{@{}c@{}}Sharma\\ et al.\cite{Sharma09} \end{tabular}} & \multicolumn{1}{|c|}{\begin{tabular}[c]{@{}c@{}}Breitenstein\\ et al.\cite{Berclaz11} \end{tabular}} & \multicolumn{1}{|c|}{\begin{tabular}[c]{@{}c@{}}Yang\\ et al. \cite{Yang09} \end{tabular}} & \multicolumn{1}{|c|}{\begin{tabular}[c]{@{}c@{}}PTracking\\ Mono \end{tabular}} & \multicolumn{1}{|c|}{\begin{tabular}[c]{@{}c@{}}PTracking\\ Multi \end{tabular}} \\ \hline
		
		\multicolumn{1}{|c|}{MOTA} & 67.0\% & 73.2\% & 67.5\% & 74.5\% & 75.9\% & \textbf{76.0}\% & \textbf{87.4}\% \\ \hline
		
		\multicolumn{1}{|c|}{MOTP} & 53.4\% & 60.3\% & 48.2\% & 56.3\% & 53.8\% & \textbf{63.0\%} & \textbf{72.2}\% \\ \hline
	\end{tabular}
	\label{tab:QuantitativeAnalysis}
\end{table*}

More specifically, KClusterize first clusters the particles trying to find all the possible Gaussian distributions. Then, a post-processing step is applied to verify that each cluster actually represents a Gaussian distribution. To this end, all the non-Gaussian clusters are split (if possible) into Gaussian clusters. It is worth noticing that, the final number of Gaussian distribution components provided as output can be different from the one found during the first step. Finally, using such clusters a GMM set $ (\boldsymbol\lambda_{s,t},\boldsymbol{\mu}_{s,t},\boldsymbol\sigma_{s,t}) $, representing the estimations performed by the camera $ s $, is created.

\begin{figure}[!t]
	\centering
	\vspace{0.1cm}
	\subfigure[]
	{
		\begin{tikzpicture}[map/.style={draw=black,ultra thick,inner sep=0pt}]
			\node at (0,0) [map]
			{
				\includegraphics[width=0.315\linewidth]{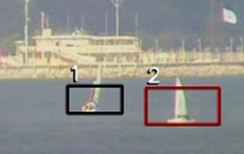}
			};
		\end{tikzpicture}
	}
	\hspace{-0.45cm}
	\subfigure[]
	{
		\begin{tikzpicture}[map/.style={draw=black,ultra thick,inner sep=0pt}]
			\node at (0,0) [map]
			{
				\includegraphics[width=0.315\linewidth]{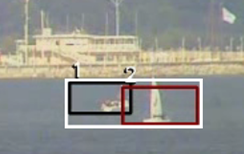}
			};
		\end{tikzpicture}
	}
	\hspace{-0.45cm}
	\subfigure[]
	{
		\begin{tikzpicture}[map/.style={draw=black,ultra thick,inner sep=0pt}]
			\node at (0,0) [map]
			{
				\includegraphics[width=0.315\linewidth]{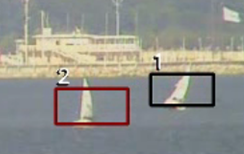}
			};
		\end{tikzpicture}
	}
	\vspace{-0.45cm}
	\caption{Group tracking. Two sailing boats are going to cross each other. Occlusions
    		 are handled considering the collapsing tracks to form a group, instead of
             tracking them separately.}
	\label{fig:GroupTracking}
\end{figure}

\textbf{Global estimation.} The global estimation phase (Algorithm \ref{alg:MCPF}, lines \ref{lab:BeginGlobalEstimation}-\ref{lab:EndGlobalEstimation}) starts receiving information from other cameras. Notice that, as already mentioned, the proposed method is asynchronous and the collection of information is limited to a small amount of time $ \Delta t $. During this time information is received from a subset $ \mathcal{S'} \subseteq \mathcal{S} $ of cameras. This mechanism is thus robust to communication delays and dead nodes, since the global estimation phase proceeds even if some node is not communicating or the communication channel is not reliable. Once data have been gathered, a particle set $ \tilde{\xi}_{\mathcal{S'},t} $ is updated using the received GMM parameters $ (\boldsymbol{i}_{s,t},\boldsymbol\lambda_{s,t},\boldsymbol{\mu}_{s,t},\boldsymbol\sigma_{s,t}) $ for $ s \in \mathcal{S'}$. These particles are re-sampled in order to extract reliable quality information about the global estimates. Then, a weighting procedure is applied to the set. Instead of weighting the particles by using the whole pool of GMM parameters, we cluster them by again using \emph{KClusterize} to obtain a new GMM pool. The weighting of particles is performed using such a new GMM pool. In this way the assigned weights are more consistent since only the most relevant parameters are considered. The global estimation phase determines the GMM parameter set of the tracked objects considering all the information available at time $ t $ (local observations and information received by other cameras). Finally, a \emph{data association} step is applied to assign an identity to each object considering all the available information received by other cameras.

\textbf{Data association.} An identity (i.e., a track number) has to be assigned to each object, by associating the new observations to the existing tracks. This is the most difficult and fundamental step for any tracking algorithm. In our approach, we consider as features for data association the direction, the velocity and the position of the objects. Complete and partial occlusions can occur when objects are aligned with respect to the camera view or when they are very close to each other, making visual tracking hard. Our solution is to consider collapsing tracks to form a group, instead of tracking them separately (see Figure \ref{fig:GroupTracking}). When multiple tracks have their bounding boxes moving closer to each other (Figure \ref{fig:GroupTracking}a), the tracker saves their color histograms and it merges them into a group (Figure \ref{fig:GroupTracking}b) - the histograms are used as models for re-identifying the objects when the occlusion phase is over (Figure \ref{fig:GroupTracking}c). A group evolves considering both the estimated trajectory and the observations coming from the detector. When an occluded object becomes visible again, the stored histograms are used to re-assign the correct identification number, belonging to the corresponding registered track.

\textbf{Quantitative analysis.} We use the CLEAR MOT \cite{Kasturi09} metrics \emph{MOTA} and \emph{MOTP} to quantitatively measure the performance of the proposed tracking method. We use the ground-truth used in \cite{Andriyenko11} and the CLEAR MOT metrics have been computed using the publicly available code provided by Zhang et al.\cite{Zhang12}. The assignment of tracking output to ground-truth is done using the Hungarian algorithm with an assignment cut-off at 1 meter. \emph{MOTP} is normalized to this cut-off threshold. Table \ref{tab:QuantitativeAnalysis} shows the quantitative comparison with state-of-the-art approaches on the PETS 2009 data set. It is worth noticing that this data set is one of the most challenging one for tracking systems. Finally, we use View1, View3 and View8 to perform the distributed tracking in the ``PTracking Multi'' setup.
\section{Experimental Evaluation}
\label{sec:Experiments}

The intention inference algorithm and the distributed tracker were tested in two different environments: Our HRI lab (see \figurename~\ref{fig:PersonYoubotScreen1}) and the main entrance to our Informatics Forum. Videos of our experiments are publicly available on our website\footnote{Videos can be downloaded from \url{http://goo.gl/r4pJIV}.}.

\subsection{Laboratory Experiments}
\label{subsec:LaboratoryExperiments}

\textbf{Setup.} Robot position and velocity estimates are acquired through \textit{adaptive Monte Carlo localization} with an on-board laser scanner per robot. Pedestrian position and velocity estimates are provided by the distributed tracker using two overhead cameras, facing opposite directions with overlapping fields of view over the environment. Each agent is delimited by a $80$ $cm^2$ circular boundary given the footprint of the robots used for the experiments. In high density navigation, autonomous robots are challenged with reacting fast enough to avoid collisions while navigating towards their goals efficiently. We use an HRVO-based fast de-centralised reactive planner for controlling our robots autonomously. 

\begin{figure}
	\centering
	\includegraphics[width=0.95\linewidth]{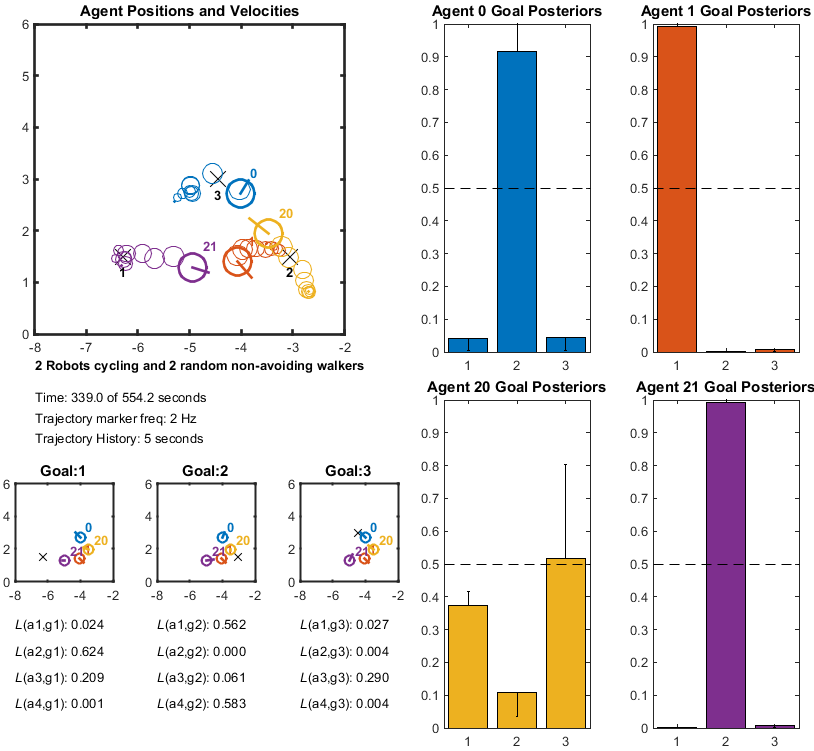}
	\vspace{-0.25cm}
	\caption{Two autonomous robots (\emph{Agents 0 and 1}) cycle clockwise and 2
			 pedestrians (\emph{Agents20 and 21}) navigate around the environment. Human
			 participants were instructed to choose random goals and to let the robots do
			 most of the avoidance. Even in complex scenarios, our goal-inference
			 algorithm provides real-time accurate intention predictions for all agents.}
	\label{fig:InSpaceData1}
\end{figure}

\textbf{Description.} Although many experiments with differing agent and task combinations were carried out, we choose to show a 4 agent navigation experiment for demonstration purposes. \figurename~ \ref{fig:InSpaceData1} shows two autonomous robots (\emph{Agents0 and 1}) tasked with moving through the goals in a clockwise cycle. Two human participants (\emph{Agents20 and 21}) randomly decide which goal to go for next after arriving at each target goal. This experiment forces both robots and humans to navigate interactively since the space for collision free motion is limited.

\textbf{Pedestrian motion.}  The accurate velocity control by the robot agents enhances the position and velocity estimates provided by the distributed tracker. People are however generally faster in both navigation speed and motion planning, representing a harder agent to track and predict. Our distributed tracker updates the agent motion parameters online and provides a representative navigation model of each agent in the environment. This enables the inference algorithm to predict human navigation goals just as fast as for autonomously planning robots.

\textbf{Performance.} During our experiments in complex scenarios including autonomous robots and human walkers, motion is fluid and convergence over posteriors occurs as quickly as $100ms$ after leaving a goal -- one single iteration of the inference algorithm. When agents are unobstructed, our algorithm performs comparable to a simpler constant-velocity model that assumes a direct trajectory towards the goal. When agents are forced to move at a velocity constrained by other agents' motion, our inference framework predicts the reciprocal change in motion accurately. Our algorithm thus converges towards the true latent goal when the observed velocity is affected by interactive constraints.
 
\figurename~\ref{fig:InSpaceData1} shows the instantaneous likelihoods and posterior estimates over goals for all agents. The inference of \emph{Agent20}'s intention is the only one not converged yet since the agent just left \emph{Goal2}. Its velocity (influenced by \emph{Agent0}'s motion) is used by our framework to predict the agent is moving towards \emph{Goal3}. Note the probability of \emph{Agent20} moving towards \emph{Goal1} is relatively high, given that it's hypothetical motion towards \emph{Goal1} could be blocked by \emph{Agent1}. During some experiments, humans were asked to not avoid the robots and navigate towards goals non interactively. Our reactive planner is still capable of evading un-cooperative agents, even though the framework is designed for fully-aware interactive navigation. Minor collisions during experiments were rare and caused due to wireless failure or complete occlusion of a camera tracked agent.

\begin{figure}[!t]
	\centering
	\includegraphics[width=0.95\linewidth]{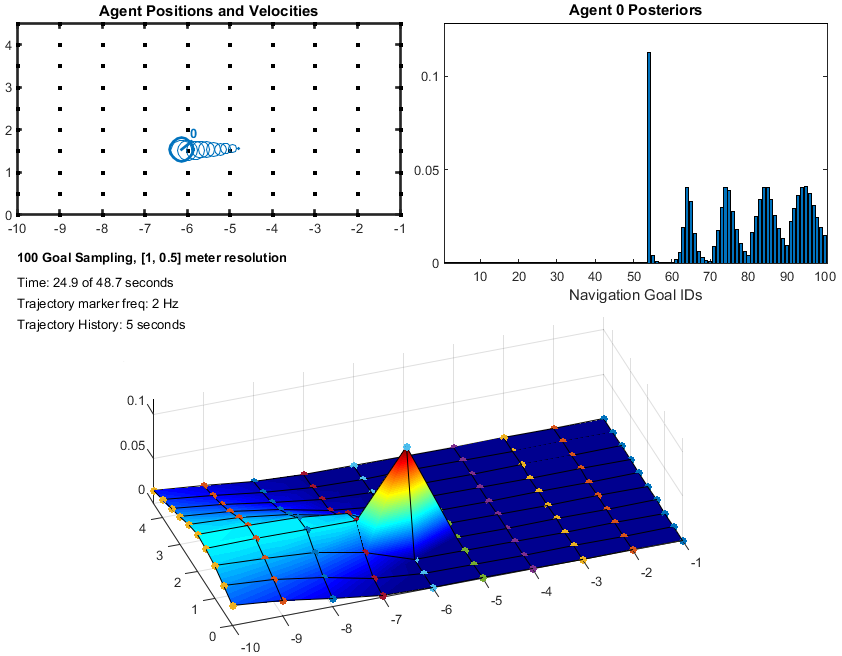}
	\vspace{-0.25cm}
	\caption{Sampling of goal space for intention inference. 100 discrete samples across
			 the x and y space dimensions at 1 and 0.5 meter separation respectively.
			 \emph{Agent0} navigates and reaches \emph{Goal1}, located at [-6.3, 1.5].
			 The 3D plot shows the probability distribution of goals over the navigation
			 space.}
	\label{fig:GoalSampling}
\end{figure}

\textbf{Goal Sampling.} Navigation goals may not be pre-defined ahead of time, such as a robot that is unaware of the human's space of goals.  For this case we may sample the space with a discrete set of goals, and use our inference algorithm to calculate the posterior probability distribution over all possible intentions. In \figurename~\ref{fig:GoalSampling}, 100 goals were placed evenly across the space, and the autonomous agent sent to navigate towards \emph{Goal1}. The plot shows that the inference framework correctly predicts the location of the agent's goal. Note the posterior distribution behind \emph{Goal1} formed by the previous motion towards \emph{Goal1} as shown by the agent trajectory. Goal sampling is specially suitable for converging over dynamic goals, such as when an agent is followed by another.

\subsection{Atrium Experiments}
\label{subsec:AtriumExperiments}

\begin{figure}[!tbp]
	\centering
	\includegraphics[width=\linewidth]{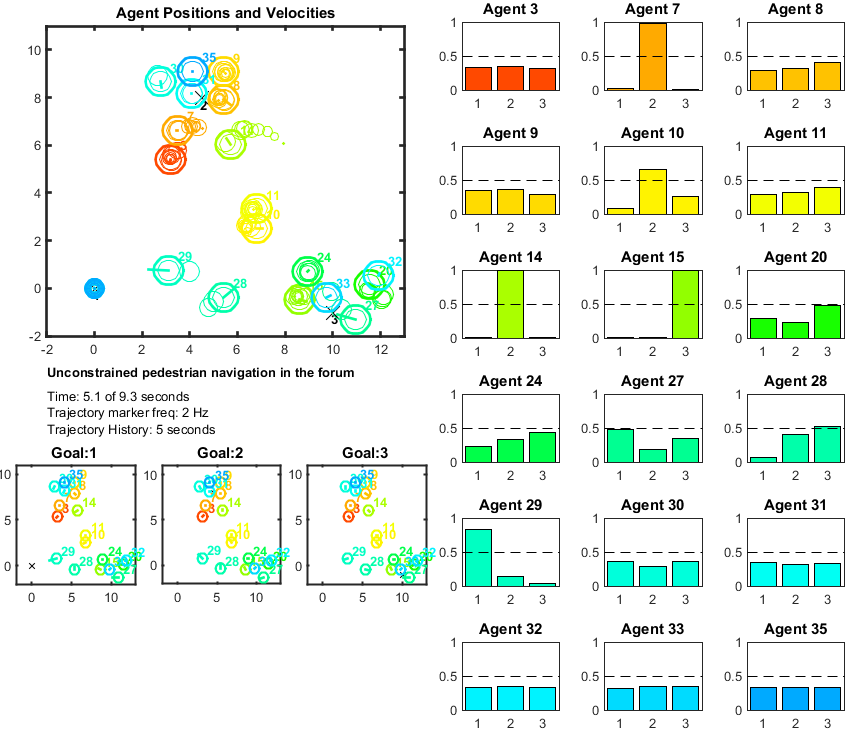}
	\vspace{-0.6cm}
	\caption{Real-time intention prediction in a densely populated environment. Around
			 20 agents navigate unconstrained in a natural scenario. In this setup, the
			 algorithm generates 60 simulated environments (20 agents, 3 goals) during
			 each inference iteration, providing an up-to-date probability distribution
			 over agent intentions.}
	\label{fig:AtriumData1}
\end{figure}

\textbf{Unconstrained.} We evaluated our framework to perform real-time tracking and goal inference in a natural human environment. This is challenging due to numerous aspects, such as containing agents with changing intentions, or navigating with other latent constraints (e.g., maintaining a formation with other agents). Our results show that, after selecting relevant goals for the environment (i.e., main exit, elevators, bathrooms), our inference algorithm provides accurate beliefs over the possible set of goals (see \figurename~\ref{fig:AtriumData1}).

\textbf{Dynamic.} The large size of this environment increases the available navigation space around agents, thus relaxing the constraint of swift collision avoidance. However, the continuous stream of agents entering and leaving the scene creates difficulties experienced by a navigating robot when navigating across a human dominated environment. Our inference algorithm is robust in dealing with any occasional identity mismatches or occlusions by the tracker.

\textbf{Density.} Given the distributed nature of our tracker and inference algorithms, computational complexity increases linearly per each agent entering the scene. This experiment shows $ \sim $20 real agents entering the environment and navigating freely between goals. Our framework is robust and goal inference accuracy remains high and convergence is fast under such a challenging setup.
\section{Methodology}
\label{sec:Methodology}

All experiments were carried out using the ROS framework. The code used for our experiments is publicly available on GitHub\footnote{PTracking can be downloaded from \url{https://github.com/fabioprev/ptracking.git} and the counterfactual framework from \url{https://github.com/ipab-rad/Youbot-RVO.git}.}.

We use a group of five KUKA YouBots in a laboratory space that covers an open space of 8 x 6 metres. The robots are autonomous, where each planner has independent knowledge and they carry out separate decision-making processes online without centralised control. Sensor fusion of data provided by the distributed tracker and robots' amcl produce accurate robot position and velocity estimates.
 
\textbf{Computability.} In order to ensure real-time performance, we measured the computational speed of our proposed method on all the environments used for the experiments. The results are produced using a single core Intel(R) Core(TM)2 Duo CPU P8400 @ 2.26GHz, 4 GB RAM. Our framework is robust at tracking, inferring and planning in real-time (Tracker: $\sim$30Hz, AMCL: $\sim$3Hz, Inference/Planner: 10Hz). Each inference step takes $\sim$3ms for a default 5 agent, 3 goal setup, scaling linearly with number of agents and goals to be inferred.
\section{Conclusions}
\label{sec:Conclusions}

We presented a novel framework for inferring and planning with respect to the movement intention of goal-oriented agents in an interactive multi-agent setup. Our counterfactual reasoning approach generates locally optimal motions of agents in the environment based on parametrised agent models, whose parameters are being estimated online from observed data. Our goal-inference procedure is a Bayesian Recursive Estimation to maintain beliefs over potential goals for all agents. This method is tested for accuracy and robustness in dense environments with autonomously planning robots and pedestrians in dynamic environments. Our results show that this is an effective and computationally efficient alternative to models that often depend on offline training of pedestrian trajectory models.
\section{Acknowledgements}
\label{sec:Acknowledgements}

This work was supported in part by grants EP/F500385/1 and  BB/F529254/1 for the University of Edinburgh School of Informatics Doctoral Training Centre in Neuroinformatics and Computational Neuroscience (www.anc.ac.uk/dtc) from the UK Engineering and Physical Sciences Research Council (EPSRC), UK Biotechnology and Biological Sciences Research Council (BBSRC), and the UK Medical Research Council (MRC).
 
\balance
\bibliographystyle{IEEEtran}
\bibliography{IEEEabrv,Main}

\end{document}